\documentclass[dvipsnames,format=sigconf,anonymous=false,review=false]{acmart}
\usepackage{natbib}
\usepackage{url}
\usepackage{color}
\usepackage{bm}
\usepackage[capitalise]{cleveref}
\usepackage{amsmath}
\usepackage{multirow}

\newcommand{\dd}{\mathrm{d}}
\DeclareMathOperator*{\argmax}{arg\,max}
\newcommand{\obs}{\mathrm{o}}
\DeclareMathOperator{\pow}{pow}

\DeclareMathOperator{\inv}{inv}

\title{Priors For Symbolic Regression}

\author{Deaglan J. Bartlett}
\affiliation{%
  \institution{CNRS \& Sorbonne Universit\'{e}, Institut d’Astrophysique de Paris (IAP),}
  \streetaddress{UMR 7095, 98 bis bd Arago,}
  \city{Paris}
  \postcode{F-75014}
  \country{France}}
\email{deaglan.bartlett@iap.fr}

\author{Harry Desmond}
\affiliation{%
  \institution{Institute of Cosmology \& Gravitation, University of Portsmouth,}
  \streetaddress{Dennis Sciama Building}
  \city{Portsmouth}
  \postcode{PO1 3FX}
  \country{UK}
}

\author{Pedro G. Ferreira}
\affiliation{%
  \institution{Astrophysics, University of Oxford,}
  \streetaddress{Denys Wilkinson Building, Keble Road}
  \city{Oxford}
  \postcode{OX1 3RH}
  \country{UK}
}

\copyrightyear{2023} 
\acmYear{2023} 
\setcopyright{acmlicensed}\acmConference[GECCO '23 Companion]{Genetic and Evolutionary Computation Conference Companion}{July 15--19, 2023}{Lisbon, Portugal}
\acmBooktitle{Genetic and Evolutionary Computation Conference Companion (GECCO '23 Companion), July 15--19, 2023, Lisbon, Portugal}
\acmPrice{15.00}
\acmDOI{10.1145/3583133.3596327}
\acmISBN{979-8-4007-0120-7/23/07}

\keywords{model selection, minimum description length, symbolic regression, equation learning, data analysis, cosmology, language model}

\begin{document}

\begin{abstract}
    When choosing between competing symbolic models for a data set, a human will naturally prefer the ``simpler'' expression or the one which more closely resembles equations previously seen in a similar context. This suggests a non-uniform prior on functions, which is, however, rarely considered within a symbolic regression (SR) framework. In this paper we develop methods to incorporate detailed prior information on both functions and their parameters into SR. Our prior on the structure of a function is based on a $n$-gram language model, which is sensitive to the arrangement of operators relative to one another in addition to the frequency of occurrence of each operator. We also develop a formalism based on the Fractional Bayes Factor to treat numerical parameter priors in such a way that models may be fairly compared though the Bayesian evidence, and explicitly compare Bayesian, Minimum Description Length and heuristic methods for model selection. We demonstrate the performance of our priors relative to literature standards on benchmarks and a real-world dataset from the field of cosmology.
\end{abstract}

\begin{CCSXML}
<ccs2012>
   <concept>
       <concept_id>10002950.10003648.10003662</concept_id>
       <concept_desc>Mathematics of computing~Probabilistic inference problems</concept_desc>
       <concept_significance>500</concept_significance>
       </concept>
   <concept>
       <concept_id>10010147.10010257</concept_id>
       <concept_desc>Computing methodologies~Machine learning</concept_desc>
       <concept_significance>500</concept_significance>
       </concept>
 </ccs2012>
\end{CCSXML}

\ccsdesc[500]{Mathematics of computing~Probabilistic inference problems}
\ccsdesc[500]{Computing methodologies~Machine learning}

\maketitle

\section{Introduction}

Symbolic regression (SR) is traditionally viewed as a Pareto optimisation problem, with the aim of finding the symbolic expressions which cannot be made more accurate without becoming more complex. The task of the SR algorithm -- whether a genetic algorithm \cite{turing, David, haupt, Lemos_2022}, a deterministic search \cite{Worm,Kammerer_2021,Dome,FFX}, reinforcement learning \cite{RL_book}, a physics-inspired search \cite{aifeyn_0, aifeyn} or an exhaustive scan of functional parameter space \citep{ESR_2022} -- is to find candidate functions, fit them to the data by minimising a loss function, and identify those that Pareto-dominate the rest. While the loss is typically straightforward to quantify as the negative of the likelihood, complexity is an inherently ambiguous notion. A variety of statistics have been constructed to quantify this (see e.g. \cite{Smits,Kommenda,Chen,Bomarito}), which may be a function of the number and/or type of operators and parameters (e.g. \cite{aifeyn_0, aifeyn,cranmer2020discovering,pysr}) or the degree of nonlinearity across the domain of interest \cite{Order_nonlinearity}.

Rather than using two separate objectives, recent work \citep{ESR_2022,Guimera_2020} has focused on combining accuracy and simplicity into a single goodness-of-fit statistic. There are fundamentally two approaches, which turn out to be closely related. 
Utilising the minimum description length (MDL) principle \cite{RISSANEN1978,MDL_review1,MDL_review2,Lanterman_2001}, \citet{ESR_2022} derived the codelength (number of nats) required to send the data with the help of a given function. By putting both on a common scale of information content, this naturally combines the preference for accuracy (lower residuals around the function's expectation) and simplicity (less information required to convey the structure of the function and its parameters).
Models are penalised which contain many terms, a large number of operators or a fine-tuning of parameters to fit the data well.
\citet{Guimera_2020} instead considered a Bayesian approach where the goodness of fit metric is the evidence. In this case the preference for functional simplicity is encoded in their prior. The choice of \citeauthor{Guimera_2020} is to assign this prior based on a corpus of equations in a given domain such that functions more similar to those in the corpus are upweighted (said to be simpler).
Each operator is assigned a probability based on its frequency in the corpus and the operators are then assumed to be independent when calculating the probability of the full function.

Our goal is to compare the Bayesian and MDL methods and propose two upgrades to the Bayesian approach, one targeting the parameter prior and one the operator prior. Our starting point in each case is the method of \citet{Guimera_2020}. They model the accuracy and parameter parts of the evidence using the Bayesian Information Criterion, an asymptotic Gaussian expansion of the likelihood that holds in the limit that the number of data points greatly exceeds the number of free parameters. 
This is however not a robust approximation to the evidence without a well-motivated prior: the posterior and hence evidence scale with the width of the prior (e.g. the range of an ``uninformative'' uniform), affecting in general the Bayes factor between two functions.
Their prior on the structure of functions treats each node of the tree independently, which is in general not desirable behaviour. For example, consider the two functions $y=\sin \left( \sin \left(x_0 + x_1 \right) \right)$ and $y=\sin (x_0) + \sin(x_1)$. Both of these functions can be expressed as trees with the same number of nodes and the same number of each type of operator, and thus a prior that depends on this would assign equal probabilities to these functions.
However, a physicist would find a sum of sines more likely \textit{a priori} than a nested sine, suggesting that
a correlation between a node and its children should be incorporated into the operator prior. We propose solutions to both problems.

The structure of the paper is as follows. We review the Bayesian and MDL model selection methods in \cref{sec:Bayes,sec:MDL} before discussing the choice of parameter prior in \cref{sec:Parameter priors}. To incorporate the correlation between operators in a prior on functions, we introduce a method for using a language model as a structural prior in \cref{sec:Function priors} and implement it using a corpus of well-known scientific
equations. We benchmark the models that we consider in \cref{sec:Benchmark} before applying them to a real-world example (the expansion history of the universe) in \cref{sec:Hubble}. We show that our new methods can lead to preference for more physically realistic equations than those in \citet{ESR_2022}. We discuss the context of our work in \cref{sec:Related work}, and conclude in \cref{sec:Conclusions}.

\section{Model selection and priors}

\subsection{Bayesian model selection}
\label{sec:Bayes}

When performing Bayesian model selection, one begins with a set of candidate models, $\{f_0, f_1, \ldots \}$, and their parameters $\{\bm{\theta_0}, \bm{\theta_1}, \ldots \}$, which aim to describe a dataset, $D$. Using Bayes' theorem, one can compute the probability of one of these models, $f_i$, given $D$ to be
\begin{equation}
    \begin{split}
        P \left(f_i | D \right) 
        &= \frac{1}{P(D)} \int 
        P \left( D | f_i, \bm{\theta}_i \right)
        P \left( \bm{\theta}_i | f_i \right)
        P \left( f_i \right)
        \dd \bm{\theta}_i \\
        &\equiv \frac{ P \left( f_i \right)}{P(D)} \mathcal{Z} \left( D | f_i \right),
    \end{split}
\end{equation}
where $P(D)$ is the probability of the data (under any model),
$\mathcal{Z} \left( D | f_i \right)$ is the evidence,
$P \left( D | f_i, \bm{\theta}_i \right)$ is the likelihood,
$P \left( \bm{\theta}_i | f_i \right)$ is the prior on the parameters,
and $P \left( f_i \right)$ is the prior on the model itself.
When comparing models on the same data one can ignore the constant $P(D)$, and thus one wishes to find the model that minimises
\begin{equation}
    \label{eq:function posterior}
    - \log P \left(f_i | D \right) = - \log P \left( f_i \right) - \log\mathcal{Z} \left( D | f_i \right).
\end{equation}

The evidence is often approximated in the case of a uni-modal posterior distribution by applying Laplace's approximation about the maximum \textit{a posteriori} point, $\hat{\bm{\theta}}_i^\mathcal{H} \equiv \argmax_{\bm{\theta}_i} \mathcal{H} (D, f_i, \bm{\theta}_i)$ for $\mathcal{H} (D, f_i, \bm{\theta}_i) \equiv P \left( D | f_i, \bm{\theta}_i \right) P \left( \bm{\theta}_i | f_i \right)$, to obtain
\begin{equation}
    \label{eq:evidence approx}
    \log\mathcal{Z} \left( D | f_i \right) \simeq
    \log \mathcal{\hat{H}}
    + \frac{p}{2} \log 2\pi 
    - \frac{1}{2}
    \log \det \hat{\bm{I}}^{\mathcal{H}},
\end{equation}
where $\mathcal{\hat{H}} = \mathcal{H} (D, f_i, \hat{\bm{\theta}}_i^\mathcal{H})$ is the maximum of the posterior, $\hat{I}^{\mathcal{H}}_{\alpha\beta} = - \partial_\alpha \partial_\beta \log \mathcal{H} (D, f_i, \bm{\theta}_i)|_{\hat{\bm{\theta}_i}^\mathcal{H}}$ and $p$ is the number of parameters of the model. In the limit of a large number of data points, $N$, the final term can be approximated by $-p/2\log N$ and the term proportional to $p$ is neglected to obtain (minus one half of) the Bayesian Information Criterion (BIC) \citep{Schwarz_1978}. Since the full evidence is typically expensive to compute, we will use the approximation given by \cref{eq:evidence approx} throughout the paper (but not the simpler BIC approximation). In all numerical experiments, $\hat{\bm{I}}^{\mathcal{H}} $ is computed via finite differences with the \textsc{numdifftools} package \cite{numdifftools}.

\subsection{Model selection by minimum description length}
\label{sec:MDL}

An alternative approach to model selection is the minimum description length (MDL) principle.
This was first applied to Genetic Programming by \citet{Iba_1994} and to SR by \citet{aifeyn_0}, and refined and formalised by \citet{ESR_2022}.
Under a two-part coding scheme, the amount of information (measured in nats, since we use natural logarithms) required to send a dataset $D$ given a hypothesis (trial function) $H$ is
\begin{equation}
    L(D) = L(H) + L(D|H),
\end{equation}
where $L(H)$ is the number of nats required to send the hypothesis, and $L(D|H)$ the number needed to send the data given this hypothesis. The optimal Shannonn--Fano coding scheme gives $L(D|H)=-\log  P \left( D | f_i, \hat{\bm{\theta}}_i^\mathcal{L} \right)$ \citep{cover_thomas} for hypothesis $H=f_i$, where $\hat{\bm{\theta}}_i^\mathcal{L} \equiv \argmax_{\bm{\theta}_i} \mathcal{L} (D, f_i, \bm{\theta}_i)$ is the maximum likelihood point.
One then chooses an encoding for the function and its parameters, $L(H)$, and selects the model that minimises $L(D)$.

\citeauthor{ESR_2022} choose this encoding such that the description length of a function comprised of 
$n$ operators,
$p$ parameters $\bm{\theta}$, with constants $\{c_\alpha\}$ and expressed as a tree with 
$k$ nodes
is
\begin{eqnarray}
    \label{eq:mdl}
    L(D) = 
    &&-\log  P \left( D | f_i, \hat{\bm{\theta}}_i^\mathcal{L} \right) + k\log(n) - \frac{p}{2} \log(3) \\ 
    && + \sum_\alpha \log(c_\alpha) + \sum_\alpha^p \left( \frac{1}{2}\log\left(\hat{I}^{\mathcal{L}}_{\alpha\alpha}\right) + \log(|\hat{\theta}_\alpha^\mathcal{L}|) \right). \nonumber
\end{eqnarray}
Note that negative constants are encoded via a sign operator that is counted in $k$ and $n$.
The term $\hat{I}^{\mathcal{L}}_{\alpha\beta} = - \partial_\alpha \partial_\beta \log P \left( D | f_i, \bm{\theta}_i \right)|_{\hat{\bm{\theta}_i}^\mathcal{L}}$ is derived by transmitting a discretised $\hat{\bm{\theta}}_\alpha^\mathcal{L}$ in the code, and considering the optimal discretisation for each of the parameters separately.

An obvious extension to Eq.~\ref{eq:mdl} is to discretise the parameters jointly rather than separately in each dimension, capturing the degeneracy between them in fitting the data. Using a rectangular lattice with a freely varying orientation in parameter space replaces $\sum_\alpha^p \left(\frac{1}{2}\log\left(\hat{I}^{\mathcal{L}}_{\alpha\alpha}\right) \right) - \frac{p}{2} \log(3)$ by $\frac{1}{2} \log \det \hat{\bm{I}}^{\mathcal{L}} + \frac{p}{2} \log(\nu_p)$, where $\log(\nu_p) = 1 - \log(3) \, \forall p$ \citep{Takeuchi_1997}. A further reduction in description length is possible by instead using an optimal quantising lattice, which gives $\log(\nu_p) = 1 + \log(4 \kappa_p)$ \citep{Wallace_1987,Wallace_1992,Lanterman_2001}. Here $\kappa_p$ is a constant relating to the geometry of the optimal lattice in $p$ dimensions, with the limits $\kappa_1 = 1/12$ and $\kappa_p \to (2 \pi e)^{-1}$ for $p \gg 1$ \citep{Conway_2013}. For simplicity we will use a rectangular lattice hereafter.

A more fundamental extension is to use prior information on the parameters to encode them optimally by assigning shorter codelengths to more \textit{a priori} probable values. In this case
the optimal parameter discretisation is determined with respect to $\hat{\bm{I}}^{\mathcal{H}}$ rather than $\hat{\bm{I}}^{\mathcal{L}}$. Combining $\log P\left( D | f_i, \bm{\theta}_i \right)$ with the parameter encoding $\log P(\bm{\theta}_i|f_i)$,
removing the $\log(|\hat{\theta}_\alpha^\mathcal{L}|)$ that derives from integer encoding and using the rectangular discretisation, the total description length becomes 
\begin{equation}
    \label{eq:mdl rectangle}
    \tilde{L} (D) = 
    -\log \mathcal{\hat{H}} +  \frac{1}{2} \log \det \bm{I}^{\mathcal{H}} + \frac{p}{2} \log(\nu_p) + k \log(n) + \sum_\alpha \log c_\alpha.
\end{equation}

By comparing to \cref{eq:function posterior} and using the approximation \cref{eq:evidence approx}, one can then identify the prior that is implicitly imposed on the functions by MDL:
\begin{equation}
    \label{eq:MDL function prior}
    - \log P \left(f_i \right) =  k \log n  + \sum_\alpha \log c_\alpha + \frac{p}{2} \log \left( 2 \pi \nu_p \right) .
\end{equation}
We see that the first two terms in this expression are also included in \cref{eq:mdl}; they penalise equations with many nodes, many different operators or many or finely tuned constants. These appear as logarithms by construction, since they are encoded using Risannen's universal prefix code for the integers \citep{Rissanen_1983}, but we have neglected the $\mathcal{O}\left(\log\log(\cdot)\right)$ term that is formally required to send all nested prefixes. This also explains the appearance of the $\log|\hat{\theta}_\alpha|$ term in \cref{eq:mdl}: without prior information on the parameters we have encoded them as integers given the precision $\Delta_\alpha$. 

The final term in the functional prior prefers functions with fewer parameters and is merely the number of these, $p$, multiplied by an $\mathcal{O}(1)$ coefficient ($\sim$0.87 for rectangular quantisation or 0.69-0.87 for optimal quantisation). These are desirable properties of a prior which prefers ``simple'' functions, making it a good choice of SR loss function in the absence of any prior preference for the type and arrangement of operators in a function. Each of the model selection metrics discussed here prefer simpler functions and hence limit or prevent overfitting, although the extent to which they do this relative to the use of a test--train split deserves further investigation.

\subsection{Priors on parameters}
\label{sec:Parameter priors}

To compute \cref{eq:function posterior}, one must choose a prior on the parameters of the function considered. The typical Bayesian view is that these encode subjective prior knowledge, but this is problematic in a SR context where one knows nothing \textit{a priori} about the thousands or millions of functions one is considering. One therefore needs an automated method for constructing the prior.

Two common choices are uniform priors on all parameters or a Jeffreys prior \citep{Jeffreys_1946} to afford reparameteristaion invariance. However, in the absence of bounds on the parameter ranges a uniform prior is improper (integrates to infinity if it is non-zero) and this is often the case for Jeffreys priors too. This makes \cref{eq:function posterior} infinite and hence unusable. Alternatively, one could introduce bounds on the parameters but this simply scales $\mathcal{Z}(D|f_i)$ and hence \cref{eq:function posterior} by an amount related to the range. Even if one could choose ``reasonable''  bounds on all parameters, this rescaling will be different for functions with different numbers of parameters and hence this method cannot provide a self-consistent model comparison metric.
The typical derivation of the BIC ignores this constant by expanding the likelihood rather than the posterior, but this appears not to be acceptable in general. One can alter the loss function of \citet{Guimera_2020} and hence the function ranking by modifying the uniform prior bounds without any change to the posteriors.
Alternative candidate priors include the reference prior \cite{Bernado_1979} and the maximum-entropy prior.
However, the reference prior, which maximises the expected information gain due to the data, reduces to the Jeffreys prior for asymptotically normal likelihoods, and hence is unsuitable as per the above. The maximum-entropy prior may include a normalisability constraint, but must include other constraints to make this feasible as the maximum-entropy distribution without any further information is an (infinite) uniform.

A solution to this problem is afforded by the \textit{Partial Bayes Factor} \citep{Lempers_1971}. Here, one splits the observed data into two samples: one for training and another for model comparison. One first runs an inference on the training sample with a given choice of prior (which does not need to be proper as the evidence is not required here) to obtain a posterior distribution of the parameters, which will be normalisable. This posterior distribution is then used as the prior for the second inference using the remaining data, from which the evidence is computed and models are compared. Of course, one must choose a size of the training set and decide which data points go into it. \citet{Berger_1996} suggest, for a chosen training set size $M$, using all possible training samples and averaging the result to obtain the \textit{Intrinsic Bayes Factor}, although which choice of average in unclear and this will require one to run 
$N! / (M! (N-M)!)$
inferences if one has $N$ data points.

To alleviate these problems, \citet{OHagan_1995} introduced the \textit{Fractional Bayes Factor} (FBF). Noting that if $M$ and $N$ are large, so that the likelihood of the test set is approximately the likelihood for the full dataset to the power $b=M/N$, \citeauthor{OHagan_1995} suggests comparing models based on
\begin{equation}
    \label{eq:FBF}
    B_b = \frac{q_1 (b)}{q_2 (b)} , \quad
    q_i (b) = \frac{ \int P \left( D | f_i, \bm{\theta}_i \right) P \left( \bm{\theta}_i | f_i \right) \dd \bm{\theta}_i }{ \int P \left( D | f_i, \bm{\theta}_i \right)^b P \left( \bm{\theta}_i | f_i \right) \dd \bm{\theta}_i},
\end{equation}
which is well-defined for an improper prior $P \left( \bm{\theta}_i | f_i \right)$.
This simply requires a choice of $b$, or test--train split fraction. The basic requirements are that $b$ must be between $M_0/N$ and 1 (where $M_0$ is the minimum permissible training set size)
and $b\to 0$ as $N \to \infty$. For robustness of training, however, one prefers a larger $b$, where ideally $N b \to \infty$ as $N \to \infty$. From these considerations, \citeauthor{OHagan_1995} suggests using $b=N^{-1}\log N$ or $b=N^{-1/2}$, where the latter is found to be more robust.
This is how models are compared in a SR context by \citep{Bomarito_2023}, where $q_i$ is computed using a Sequential Monte Carlo algorithm. One could also vary $b$ to find the corresponding uncertainty on $q_i$, but for simplicity we too choose $b=N^{-1/2}$.

To apply the Fractional Bayes Factor to the Bayesian or MDL model selection procedure, we note that \cref{eq:FBF} implies that one can take all previous formulae and replace $P(\bm{\theta}_i|f_i)$ by $P_b(\bm{\theta}_i|f_i)$, where
\begin{equation}
    \label{eq:FPF prior}
    P_b\left( \bm{\theta}_i|f_i \right) = 
    \frac{ P \left( \bm{\theta}_i | f_i \right)}{ \int P \left( D | f_i, \bm{\theta}_i^\prime \right)^b P \left( \bm{\theta}_i^\prime | f_i \right) \dd \bm{\theta}_i^\prime}.
\end{equation}
\sloppy In a SR context where this is applied to a great number of equations, one may want to approximate \cref{eq:FPF prior} for rapid computation. Under the Laplace approximation, expanding about $\tilde{\bm{\theta}}_i \equiv \argmax_{\bm{\theta}_i} \mathcal{H}_b (D, f_i, \bm{\theta}_i)$, where we define
$\mathcal{H}_b (D, f_i, \bm{\theta}_i) \equiv P \left( D | f_i, \bm{\theta}_i \right)^b P \left( \bm{\theta}_i | f_i \right)$, \cref{eq:FPF prior} becomes
\begin{equation}
    \label{eq:FPF prior approx}
    \log P_b\left( \bm{\theta}_i|f_i \right) \simeq
    \log \frac{P \left( \bm{\theta}_i | f_i \right)}{\mathcal{H}_b (D, f_i, \tilde{\bm{\theta}}_i)}
    - \frac{p}{2} \log 2\pi 
    + \frac{1}{2} \log \det \tilde{\bm{I}}^{\mathcal{H}_b},
\end{equation}
where
$\tilde{I}^{\mathcal{H}_b}_{\alpha\beta} = - \partial_\alpha \partial_\beta \log \mathcal{H}_b (D, f_i, \bm{\theta}_i)|_{\tilde{\bm{\theta}_i}}$. To compute the description length (\cref{eq:mdl rectangle}) or the approximation to the evidence in \cref{eq:evidence approx}, one must in general perform two parameter optimisations: first, one must find $\tilde{\bm{\theta}}_i$, then use this value to compute $\hat{\bm{\theta}}_i$ under the new prior, where $\hat{\bm{\theta}}_i = \argmax_{\bm{\theta}_i} P \left( D | f_i, \bm{\theta}_i \right) P_b\left( \bm{\theta}_i|f_i \right)$.

In this framework, wide and uniform priors, $P(\bm{\theta}_i|f_i) = {\rm const}$, are acceptable and even desirable, so we adopt them. In this case, $\hat{\bm{\theta}}_i = \tilde{\bm{\theta}}_i$ and these are equal to the maximum likelihood value, so this is a special case where one only needs to perform a single optimisation (of the likelihood).
One also sees that $b^{-1} \tilde{I}^{\mathcal{H}_b}_{\alpha\beta} = \tilde{I}^{\mathcal{H}}_{\alpha\beta} = \tilde{I}^{\mathcal{L}}_{\alpha\beta}$,
so 
$ b^{-p} \det \tilde{I}^{\mathcal{H}_b} = \det \tilde{I}^{\mathcal{H}} = \det \tilde{I}^{\mathcal{L}}$. Subsituting these results into \cref{eq:FPF prior approx,eq:evidence approx}, one finds for a FBF prior with uniform $P(\bm{\theta}_i|f_i)$,
\begin{equation}
    \label{eq:FBF evidence approx}
    \log\mathcal{Z} \left( D | f_i \right) \simeq
    (1-b) \log P \left(D | f_i, \hat{\bm{\theta}}_i \right) +
    \frac{p}{2} \log b.
\end{equation}
Note that for $b=N^{-1/2}$ this is almost proportional to the BIC, but with an extra factor of $1/2$ multiplying the $p\log N$ term and an extra term  $N^{-1/2} \log P \left(D | f_i, \hat{\bm{\theta}}_i \right)$, which becomes unimportant for large $N$. It is interesting that when using a uniform training-set prior any dependence on $I^\mathcal{L}$ has cancelled upon application of the FBF, independent of the choice of $b$.
This may be undesirable from a model selection perspective compared to the prior on the integers used by \citet{ESR_2022}, since it does not penalise finely tuned parameters.

\subsection{Priors on functions}
\label{sec:Function priors}

We now turn to construction of an optimal prior on the structure of candidate functions.

\subsubsection{Formalism}

Suppose we have a function represented as a tree, $\bm{T}$, and define $\bm{\mathcal{D}}_i$ to be the set of nodes of depth $i$, i.e. $\bm{\mathcal{D}}_0$ is the root node, $\bm{\mathcal{D}}_{i}$ contains the nodes of the children of $\bm{\mathcal{D}}_{i-1}$ and $i = 0, \ldots, d-1$. By the chain rule, the probability of this tree is
\begin{equation}
    \label{eq:prob tree exact}
    P \left( \bm{T} \right) = \prod_{i=0}^{d-1} P \left( \bm{\mathcal{D}}_i | \bm{\mathcal{D}}_{i-1}, \ldots, \bm{\mathcal{D}}_0 \right),
\end{equation}
where $P \left( \bm{\mathcal{D}}_i | \bm{\mathcal{D}}_{i-1}, \ldots, \bm{\mathcal{D}}_0 \right)$ is the conditional probability of obtaining $\bm{\mathcal{D}}_i$ given all the nodes with smaller depths. Our goal is to learn this probability from a set of equations in order to produce a function prior that encapsulates domain-specific knowledge concerning the types of equations one expects to find. We achieve this by training a language model on the equation set to assign probabilities to operators based on those that precede it in the tree.

We begin with two simplifying assumptions:
\begin{enumerate}
    \item We approximate \cref{eq:prob tree exact} as an $n$-gram, such that the probability of a set of nodes with given depth is only dependent on nodes which have a depth within some value, $n$.
    \item We group all nodes at a given depth into siblings of nodes which share a parent ($\bm{\mathcal{D}}_i = \{ s_i^0, \ldots, s_i^{n_i -1 } \}$ where $s_i^j$ is the the $j$\textsuperscript{th} set of siblings for $\bm{\mathcal{D}}_i$). We assume that the probability of each set of siblings only depends on their ascendants and that all sets of siblings are independent at a given depth.
\end{enumerate}
Under these approximations, one can write
\begin{equation}
    \label{eq:prob tree approx}
    P \left( \bm{T} \right) \approx
    \prod_{i=0}^{d-1} \prod_{j=0}^{n_i - 1} P \left( s_i^j | a_{i-1}^{ij}, \ldots a_{i-(n-1)}^{ij} \right),
\end{equation}
where $a_{k}^{ij}$ is the direct ancestor of siblings $s_i^{j}$ at depth $k$.

To evaluate the conditional probabilities of the $n$-grams we use a Katz back-off model \citep{Katz_1987}, although we note that our approach can easily be extended to other language models. We choose this model for its simplicity given its ability to assign a non-zero probability to expressions not in the training set. We are therefore not simply searching for equations which have appeared in the corpus, but rather are upweighting equations more similar to those in the counts and ordering of their operators.

For this model, the conditional probability of the word (operator) $w_i$ following the sequence of words (phrase) $w_{i-n+1}\ldots w_{i-1}$ is 
\begin{equation}
    \begin{split}
        &P \left(w_i | w_{i-n+1}, \ldots, w_{i-1} \right) \\
        &= 
        \begin{cases}
	   \mbox{$d_{w_{i-n+1}, \ldots, w_{i}} \frac{C(w_{i-n+1}, \ldots, w_{i})}{C(w_{i-n+1}, \ldots, w_{i-1})}$} & C(w_{i-n+1}, \ldots, w_{i}) > k \\
	   \mbox{$\alpha_{w_{i-n+1}, \ldots, w_{i-1}} P\left(w_i | w_{i-n+2}, \ldots, w_{i-1} \right)$} & \text{otherwise},
        \end{cases}
    \end{split}
\end{equation}
where $C(\cdot)$ is the number of times a given phrase has been seen in training.
$k$ is a hyperparameter determining how many times one needs to see a phrase before using the $n$-gram.
If a phrase appears fewer than $k$ times the algorithm ``backs off'' to the $(n-1)$-gram to evaluate the probability.
We choose $k=0$ for simplicity.
We compute the amount of discounting, $d_{(\cdot)}$, to be $C^\ast(\cdot) / C(\cdot)$, where $C^\ast(\cdot)$ is the Good-Turing estimate \citep{Good_1953,Gale_1995} of the number of counts of a phrase.
The left-over probability mass for the $(n-1)$-gram is
\begin{equation}
    \begin{split}
        &\beta_{w_{i-n+1}, \ldots, w_{i-1}} \\
        &= 1 - 
        \sum_{w_i : C(w_{i-n+1}, \ldots, w_{i}) > k}
        d_{w_{i-n+1}, \ldots, w_{i}}
        \frac{C(w_{i-n+1}, \ldots, w_{i})}{C(w_{i-n+1}, \ldots, w_{i-1})},
    \end{split}
\end{equation}
and thus the back-off weight is defined to be
\begin{equation}
    \alpha_{w_{i-n+1}, \ldots, w_{i-1}}
    =
    \frac{\beta_{w_{i-n+1}, \ldots, w_{i-1}}}{\sum_{w_i : C(w_{i-n+1}, \ldots, w_{i}) \leq k} P\left(w_i | w_{i-n+2}, \ldots, w_{i-1} \right) }.
\end{equation}
If no data are available to evaluate these formulae, then one recursively considers the next-shortest $n$-grams until such data exists.

For any function (whether from the corpus or if we are evaluating the prior for this function), we first read the mathematical expression using the \textsc{sympy} package \citep{Sympy} and convert this to a function tree using \textsc{esr} \citep{ESR_2022}. We then identify all sets of siblings and the direct ancestors of these siblings, which are the ``phrases'' contained in this function. The set of words used for this phrase are then the set of ancestor nodes and the terminating sibling. We split each sibling into a left and right node, and multiply the probability of obtaining the left node by the conditional probability of obtaining the right node given the left one. Since the structure of the phrases are different for these two probabilities (one always ends with a single leaf node, but the other has two leaf nodes), we use a different back-off model for each and multiply the results.

\subsubsection{Example - Scientific equations}
\label{sec:Science equations}

For the applications of the language model prior considered in this work, we use a compilation of 161 scientific equations as our training data. This contains the 100 equations in the Feynman Symbolic Regression Database (FSReD) \citep{aifeyn_0}, which were selected from the \textit{Feynman Lectures on Physics} \citep{feynman_v1,feynman_v2,feynman_v3}. We also include the 20 ``bonus'' equations from this database, which were taken from other physics textbooks \citep{Goldstein_2002_classical,Jackson_1999_classical,Weinberg_1972,Schwartz_2014_quantum}.
Finally, inspired by \citet{Guimera_2020}, we augment our training set with a selection of 41 equations taken from pages linked to Wikipedia's ``List of scientific equations named after people.''
These cover equations found in topics such as classical
mechanics, electromagnetism and special relativity, contain the operators
$\{+$,
$-$,
$\times$,
$\div$,
$\pow$,
$\sqrt{\cdot}$,
$\exp$,
$\log$,
$\sin$,
$\cos$,
$\arcsin$,
$\arccos$,
$\tanh\}$
and are functions of between one and eleven variables.

\begin{figure}
    \centering
    \includegraphics[width=\columnwidth]{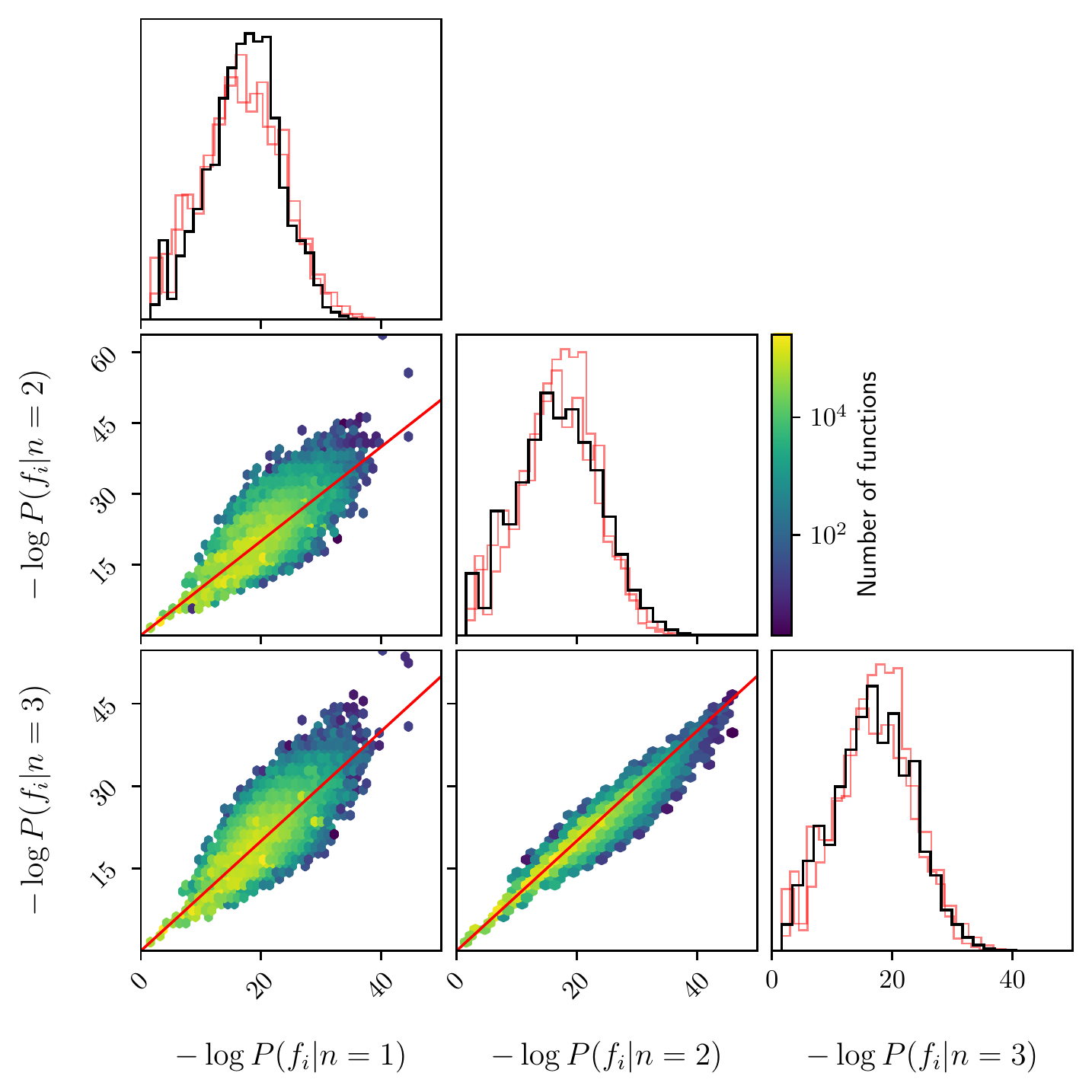}
    \caption{
    One- and two-dimensional histograms of prior probabilities for all functions up to complexity 10 using the basis
    $\{x$,
    $a$,
    $\inv$,
    $+$,
    $-$,
    $\times$,
    $\div$,
    $\pow \}$.
    The priors are based on the compilation of scientific equations described in \cref{sec:Science equations}
    and are evaluated using a back-off model with different lengths, $1\le n\le 3$. For the two-dimensional plots, points lying on the red line have an equal prior for both values of $n$ in the corresponding panel. 
    In the one-dimensional plots, the black lines are the distributions for the corresponding $n$ and we plot the histograms for the other values of $n$ in red for reference.
    }
    \label{fig:backoff_length}
\end{figure}

The remaining degree of freedom is the length of the $n$-grams used to produce prior probabilities. In \cref{fig:backoff_length} we investigate the effect of varying $n$ by constructing a back-off model with the FSReD functions with different values. We evaluate the prior probabilities of all equations which can be constructed from the basis
$\{x$,
$a$,
$\inv$,
$+$,
$-$,
$\times$,
$\div$,
$\pow \}$
up to complexity (number of tree nodes) 10. Here $n=1$ means that we do not consider the effect of any parents, for $n=2$ the probabilities are conditioned on the parent nodes only, and for $n=3$ we condition on both the parent and the grand-parent nodes.
From the large scatter when comparing $n=1$ and $n=2$ we learn that there is significant information in the parent nodes. This is expected (given a node, some children are more likely than others), illustrating the importance of modelling correlations in the function prior. There is, however, less scatter in the probabilities between $n=2$ and $n=3$, hence we choose $n=2$. We find that the one-dimensional distributions are very similar for all $n$.

\captionsetup{belowskip=-10pt}
\captionsetup{belowskip=0pt}

\section{Benchmark tests}
\label{sec:Benchmark}

We have discussed a range of methods for SR model selection from either a Bayesian or MDL perspective, involving different choices of prior or using more heuristic approaches. To study the effect of these choices we apply the following six model selection procedures to a set of benchmark datasets:
\begin{enumerate}
    \item \textit{Likelihood:} Choose the model with the maximum likelihood, $P(D|\bm{\theta}_i, f_i)$.
    \item \textit{Score:} As a heuristic approach, choose the model which has the best ``score'', defined to be the model on the Pareto front with the largest negative of the derivative of log-likelihood with respect to complexity (number of nodes in the tree representation). This is the choice adopted by \textsc{PySR} \cite{cranmer2020discovering,pysr}. Note we always report the logarithm of the score due to the potentially large range of values.
    \item \textit{MDL:} Choose the MDL model using the prescription of \citet{ESR_2022} (\cref{eq:mdl}).
    \item \textit{MDL+LM:} Choose the MDL model according to \cref{eq:mdl}, but using the prior on functions described in \cref{sec:Function priors} instead of the term $k\log n$.
    \item \textit{MDL+FBF+LM:} Choose the MDL model using FBF priors on initially uniform parameter priors and with a language model function prior.
    \item \textit{Bayes+FBF+LM:} Choose the model with the largest Bayesian probability (\cref{eq:function posterior}) using the approximation to the Bayesian evidence from a FBF procedure with initially uniform parameter priors (\cref{eq:FBF evidence approx}) and using the language model prior discussed in \cref{sec:Function priors}.
\end{enumerate}
We choose these methods since they encompass a broad range of model selection approaches, however we leave a more comprehensive survey of different methods to future work.

To benchmark these methods, we consider the 
five
standard test functions given in \cref{tab:bench_fun} which were proposed as SR benchmarks by \citet{Nguyen_2011} and \citet{Korns_2011} and collated by \citet{McDermott_2012}, where we use domains comparable to these references. Since our model selection methods are designed for noisy data, we first generate $10^5$ values of $x$ from a uniform distribution and measure the standard deviation of the resulting $y$ values. For all our benchmark studies, we add Gaussian noise with a standard deviation, $\sigma$,
equal to half this value.
Our likelihood is then
\begin{equation}
        \log P \left( D | f_i, \bm{\theta}_i \right) = 
        -\frac{1}{2 \sigma^2}
        \sum_a \left(y_{a} - f_i(x_a|\bm{\theta}_i) \right)^2 + {\rm const},
\end{equation}
where we sum over the mock data points $\{(x_a, y_a)\}$.
To ensure our results are robust to the random seed, for each function and choice of number of data points, we generate five data sets which we independently analyse.

Since our focus here is on model selection rather than equation generation we use the \textsc{esr} algorithm so that we are guaranteed to evaluate all functions of a given complexity, rather than a subset as would be generated using a non-exhaustive method. We use the operator basis set
$\{ x$,
$a$,
$\sqrt{\cdot}$, 
$(\cdot)^2$,
$\sin$,
$\cos$,
$+$,
$\times$,
$-$,
$\div$,
$\pow \}$
and run \textsc{esr} to complexity 8.
Although one of the benchmark functions contains an exponential, we do not include this operator since the generating function has a lower-complexity variant using the power operator.
We stress, however, that these model selection procedures do not require an exhaustive search so could equally well be implemented in conjunction with any other function generation algorithm.

\begin{figure*}
    \centering
    \includegraphics[width=\textwidth]{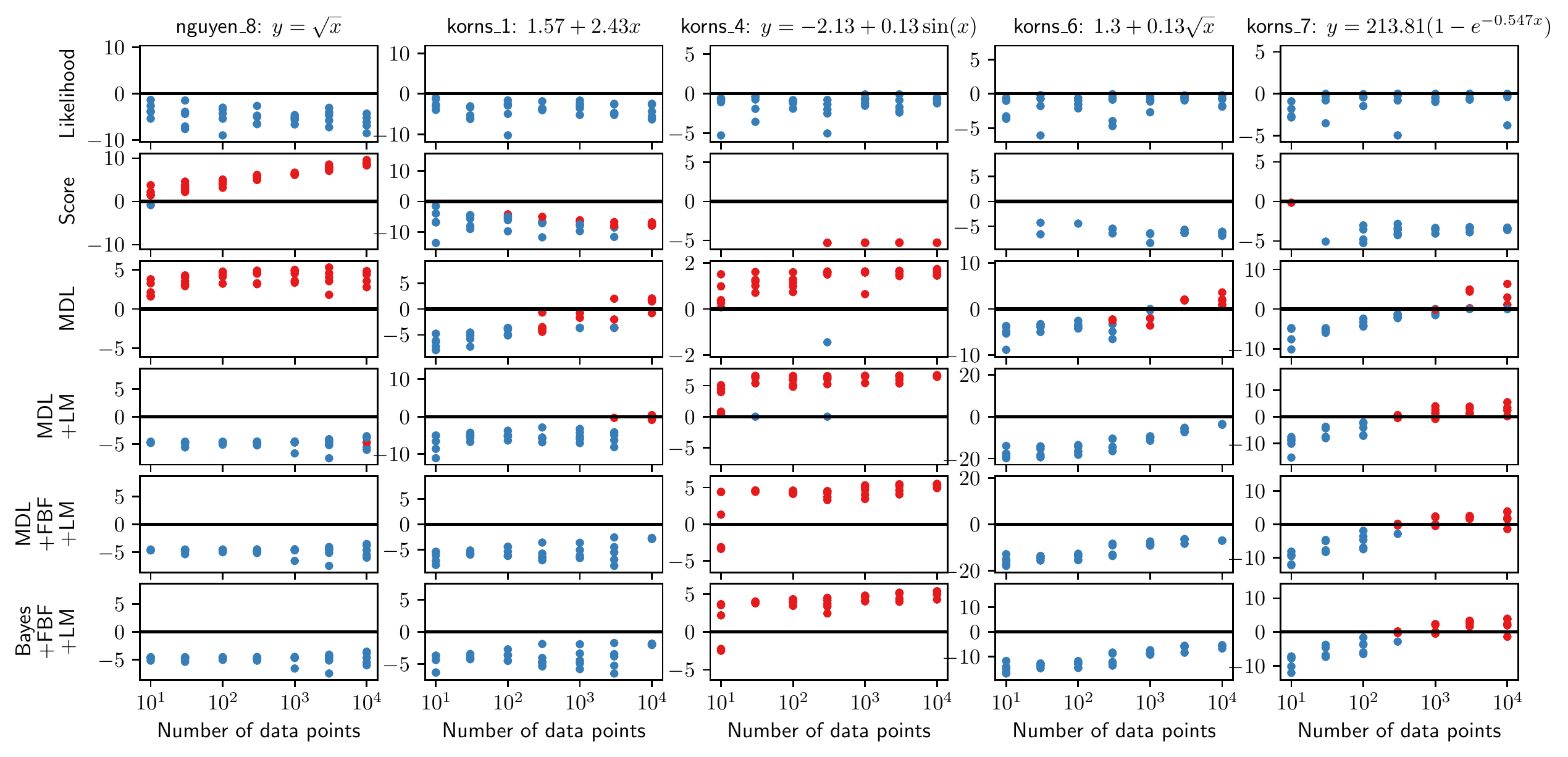}
    \caption{Preference for the truth for benchmark tests according to different model-selection methods as a function of number of data points, when considering functions up to complexity 8 with the basis
    $\{ x$,
    $a$,
    $\sqrt{\cdot}$, 
    $(\cdot)^2$,
    $\sin$,
    $\cos$,
    $+$,
    $\times$,
    $-$,
    $\div$,
    $\pow \}$.
    Any point above zero favours the truth. Red points indicate runs where the truth is in the top two functions, otherwise the point is blue.
    }
    \label{fig:benchmark_results}
\end{figure*}

Our results are plotted in \cref{fig:benchmark_results}, where we show the difference in each metric between the top-ranked function and the truth (if the truth is not ranked top) or the truth and the second-best function (if the truth is ranked top).
We consider the ``truth'' to be any function mathematically equivalent to the generating function as given in \cref{tab:bench_fun}. This is determined by the function simplification step of \textsc{esr} and subsequent visual inspection.
The model selection method has successfully recovered the truth for all points above zero in the plot, and the further above zero the stronger the preference for the truth compared to any other function. If the truth lies in the top two functions, we colour the point red, otherwise it is blue. 

One sees that assessing functions solely based on their likelihood does not find the truth to be in the top two equations for any of the cases considered. This is as one would expect for noisy data: one can make a function arbitrarily accurate by making it sufficiently complex, and thus the relatively simple generating equations are not recovered and over-fitting is preferred. The `score' method performs better, almost always selecting the truth for the Nguyen-8 function, although in many other cases the truth does not appear anywhere in the function ranking. This is a consequence of the score only being defined on the Pareto-front; if the truth does not lie on this (even if it is infinitesimally far away), then this metric cannot accommodate it. Perhaps more concerning is that, even if the truth is on the Pareto-front, this metric does not necessarily prefer the truth as the number of data points tends to infinity; the likelihood scales approximately linearly with the number of data points, and thus the overall ranking does not necessarily change.

Turning our attention to the MDL and Bayesian metrics, we find that these more often select the truth than the other methods. For the simplest function (Nguyen-8), the MDL prescription of \citet{ESR_2022} performs best and always selects the truth. We find that the methods involving the language model prior do not perform as well for this equation, with the truth often appearing as the third best function. However, we note that $y=x^{\theta_0}$ is frequently the second-best function in this case, which is a generalisation of the truth. This is likely a consequence of variables being raised to a power being more common in our training set than square-roots. 
For the Korns-1 and Korns-6 functions, the MDL approach outperforms the other methods, although larger sample sizes are required to recover the truth than for the Nguyen-8 data. The MDL+LM method is the only other procedure which can select the truth for any run considered, correctly identifying the generating equation for Korns-1 when $\sim10^4$ data points are used.
All MDL and Bayesian methods perform similarly well for the Korns-4 dataset, with the truth almost always being selected, contrary to the other approaches.

The language model methods outperform the other criteria when applied to the Korns-7 dataset. Since the function prior is based on scientific equations and we are applying this to a different context (namely functions used as SR benchmarks), there is no reason to suppose \textit{a priori} that this would be the case. However, when one examines the truth, one can imagine a plethora of scientific applications for such a function. As such, including a prior based on these equations allows the truth to be selected with fewer data points than would be required otherwise. 

\begin{table}
    \begin{center}
		\begin{tabular}{|c|c|c|c|}
		\hline
            Dataset & Function & Complexity & Domain \\
		\hline
            Nguyen-8 & $\sqrt{x}$ & 2 & $[0,4]$ \\
            Korns-1 & $1.57 + 2.43 x$ & 5 & $[-50, 50]$\\
            Korns-4 & $-2.13 + 0.13 \sin(x)$ & 6 & $[-50, 50]$ \\
            Korns-6 & $1.3 + 0.13 \sqrt{x}$ & 6 & $[0, 50]$ \\
            Korns-7 & $213.81 (1 - e^{-0.547 x})$ & 7 & $[0, 50]$\\
            \hline
		\end{tabular}
    \caption{Datasets used as benchmarks, taken from \citep{Nguyen_2011} (Nguyen) and \citep{Korns_2011}  (Korns), as collated in \citep{McDermott_2012}.}
    \label{tab:bench_fun}
    \end{center}
\end{table}

\section{Physically reasonable functions}
\label{sec:Hubble}

We now apply our language model prior to the context for which it is designed: to investigate how it affects the ranking of functions we apply it to a real-world SR problem drawn from the field of cosmology. We reanalyse the results of \citet{ESR_2022} where the expansion rate of the Universe, $H$, was determined as a function of the redshift, $z$ (the fractional change in a photon's wavelength due to the expansion of the Universe; a higher redshift is an earlier time in the history of the Universe) using the \textsc{esr} algorithm. This is a univariate SR problem, where one wishes to find $y \left(x \equiv 1 + z \right) \equiv \left( H\left( z \right) \right)^2$, which, according to the concordance cosmological model (called $\rm \Lambda CDM$), satisfies the Friedmann equation
\begin{equation}
    \label{eq:Friedmann_LCDM}
    y_{\rm \Lambda CDM} \left(x \right) = \theta_0 +  \theta_1 x^3.
\end{equation}
We note that this equation does not appear in our corpus, so we expect to `discover' a new equation in this case study.

In particular we focus on the analysis of the Pantheon+ sample of 1590 Cepheid-calibrated supernovae \cite{Scolnic_2021,SH0ES_2022}. The observable here is not $y$, but the ``distance modulus'' $\mu$ which can be calculated by computing an integral of $y$ (see \citep{ESR_2022}). The likelihood in this case is a modelled as a multivariate Gaussian
\begin{equation}
    \begin{split}
        \log P \left( D | f_i, \bm{\theta}_i \right) = 
        &-\frac{1}{2}
        \left(\bm{\mu}_i \left( \bm{z}^\obs \right) - \bm{\mu}^\obs  \right)^{\rm T}
        \mathbf{\Sigma}^{-1}
        \left(\bm{\mu}_i \left( \bm{z}^\obs \right) - \bm{\mu}^\obs \right) \\
        & - \frac{1}{2}\log \left( \det \left( 2 \pi \mathbf{\Sigma} \right) \right),
    \end{split}
\end{equation}
where we use the publicly available covariance matrix, $\bm{\Sigma}$ \citep{Pantheon_2022}, $\bm{\mu}_i$ is the vector of predicted distance moduli at redshifts $\bm{z}^\obs$ using equation $f_i$ with parameters $\bm{\theta}_i$, and $\bm{\mu}^\obs$ are the observed values.

In the absence of a prior on functions from the literature, the functions were ranked according to their description length (\cref{{eq:mdl}}). Although \textsc{esr} was run up to complexity 10, the MDL function was found to be
\begin{equation}
    y_{\rm MDL} \left( x \right) = \theta_0 x^x = \theta_0 \left(1 + z \right)^{\left( 1 + z \right)},
\end{equation}
which has a complexity of 5. With very few nodes in the tree representation and a small number of operators, it is clear that such a function would have a small penalisation due to model complexity in the MDL framework. However, $x^x$ rarely appears in empirical science and one would unlikely expect \textit{a priori} for an equation containing it to describes the expansion of the Universe.

We therefore re-evaluate the functions considered by \citeauthor{ESR_2022} but using the six methods outlined in \cref{sec:Benchmark}. We give the top-4 functions for each method in \cref{tab:sn_results}, keeping the highest-ranked variant for equivalent expressions and combining functions of parameters into new parameters during post-processing. The functions given in \cref{tab:sn_results} all produce reasonable likelihoods, with a mean Mahalanobis distance
($N^{-1} \left(\bm{\mu}_i \left( \bm{z}^\obs \right) - \bm{\mu}^\obs  \right)^{\rm T}
\mathbf{\Sigma}^{-1}
\left(\bm{\mu}_i \left( \bm{z}^\obs \right) - \bm{\mu}^\obs \right)$ for $N=1590$)
between 0.8783 and 0.9223.

When one only replaces the $k\log n$ term of \cref{eq:mdl} with the language model prior, one sees that the physically unreasonable $x^x$ functions are no longer in the top-four. Instead, all of these equations are merely sums of $x$ to a (positive or negative) integer power, with at least one free parameter to set the scale of the expansion rate. Moreover, \cref{eq:Friedmann_LCDM} (which has a strong theoretical prior based on General Relativity) is ranked much higher (fourth) with this method. This suggests that the language model, as expected, is helping to eliminate the physically unreasonable functions.
Interestingly, we see that the language model often selects functions which divide by parameters, rather than multiply. In the absence of a language model prior, this can occur due to the different Fisher terms of $\theta_0$ and $1/\theta_0$ in \cref{eq:mdl}, however we note that our function prior also often prefers division. For example, the function ${x (2 x + 1/{x})}/{\theta_{0}}$ is preferred to ${x (2 x + 1/{x})}\times{\theta_{0}}$ by a change in the log-prior of 0.41. Choosing a different corpus of equations could reverse this preference,
highlighting the importance of using an appropriate prior for the given context.

Turning our attention to the criteria involving the FBF, the results are more mixed. Although the $x^x$ behaviour is removed, the highest ranked equation for both methods is still somewhat unappealing. Each of the constituent elements of the function are however reasonable and thus this function's high ranking may be due to the low number of consecutive operators considered by the back-off model. Alternatively it could be due to a well known problem with Katz back-off models, that if a phrase has not been seen before but is comprised of two often-encountered sub-phrases then it will necessarily be assigned a high probability \citep{Manning_1999}. Thus this function may be down-weighted with a more sophisticated language model. The remaining functions all appear reasonable, and the Bayesian method finds \cref{eq:Friedmann_LCDM} in the top-four equations.

\begin{table*}
    \begin{center}
		\begin{tabular}{|c|c|c|c|c|c|c|c|}
		\hline
            \multirow{1}{*}{Method} & \multicolumn{1}{c|}{Rank 1} & \multicolumn{2}{c|}{Rank 2} & \multicolumn{2}{c|}{Rank 3} & \multicolumn{2}{c|}{Rank 4} \\
		\cline{2-8}
		& Function & Function & Value & Function & Value & Function & Value \\
		\hline
Likelihood & $\left|{\theta_{1} + \frac{1}{\theta_{0} - x}}\right|^{\left|{\theta_{2}}\right|^{x}}$ & $\left|{\theta_{0} - x^{\theta_{1}}}\right|^{\theta_{2} - x}$ & 1.48  & $- \theta_{2} + \left|{\theta_{0} \left(\theta_{1} - x\right)}\right|^{x}$ & 3.15 & $\frac{\left|{\theta_{1} + x^{\theta_{2}}}\right|^{x}}{\theta_{0}}$ & 3.18\\
\hline
Score &  $\theta_{0} x$ & $\theta_{0} \left|{\theta_{1}}\right|^{- x}$ & 11.54 & $\left|{\theta_{0} - x^{\theta_{1}}}\right|^{\theta_{2} - x}$ & 14.32 & $\left|{\theta_{1} + \frac{1}{\theta_{0} - x}}\right|^{\left|{\theta_{2}}\right|^{x}}$ & 14.59 \\
\hline
MDL & $\theta_{0} x^{x}$ & $\left|{\theta_{0}}\right|^{x^{\theta_{1}}}$ & 0.78 & $\theta_{0} \left|{\theta_{1}}\right|^{- x}$ & 0.83 & $\theta_{0} x^{x^{\theta_{1}}}$ & 0.91 \\
\hline
MDL+LM & $\frac{x \left(2 x + \frac{1}{x}\right)}{\theta_{0}}$ & $\frac{2 x^{2} + \frac{1}{x}}{\theta_{0}}$ & 1.40 & $\frac{x^{2} + 2 x}{\theta_{0}}$ & 1.63 & $\theta_{0} + \theta_{1} x^{3}$ & 1.72 \\
\hline
MDL+FBF+LM & $\left|{\theta_{0}}\right|^{\frac{1}{\theta_{1} + 2 x}}$ & $\left|{\theta_{0}}\right|^{\left|\theta_1\right|^{x}}$ & 0.16 & $\theta_{0} \left|{\theta_{1}}\right|^{x}$ & 1.77 & $\frac{\theta_{1} - x^{2}}{\theta_{0}}$ & 2.19\\
\hline
Bayes+FBF+LM & $\left|{\theta_{0}}\right|^{\frac{1}{\theta_{1} + 2 x}}$ & $\left|{\theta_{0}}\right|^{\left|\theta_1\right|^{x}}$ & 0.16 & $\left|{\theta_{0}}\right|^{- \theta_{1} + x}$ & 1.03 & $\frac{x^{3} + \frac{1}{\theta_{0}}}{\theta_{1}}$ & 1.95  \\
\hline
		\end{tabular}
    \caption{Four highest ranked functions for $y(x\equiv1+z)=H^2(z)$ for the Pantheon+ supernova sample according to different model selection methods. Here, MDL, FBF and LM mean Minimum Description Length, Fractional Bayes Factor and Language Model, respectively (see text for details). The values of each metric are given relative to the highest-ranked function.}
    \label{tab:sn_results}
    \end{center}
\end{table*}

\section{Related work}
\label{sec:Related work}

The idea of using transformers as a proposal strategy for generating candidate equations given a set of observed input-output pairs has been developed before \citep{Valipour_2021,Kamienny_2022,Biggio_2021,Shojaee_2023,Vastl_2022,dAscoli_2022,Becker_2022}. 
Typically, the transformers are trained on several million functions, which are generated randomly by first producing candidate trees (see e.g. \citep{Lample_2019}) and then decorating these with operators from a given basis set. The range of input variables and parameters of the functions are then drawn from some random distribution and the function evaluated. Once trained, the transformer takes as input the observed inputs and outputs and suggests functions which could have generated the data. The constants of these equations can then be fine tuned by a further optimisation step. One can interface the decoder with a Monte Carlo Tree Search to provide feedback during the generation process to help select equations which optimise some metric of accuracy and/or complexity \citep{Shojaee_2023}. Similarly, one can use a probabilistic grammar as a generative model \citep{Brence_2021}, such that different productions have different probabilities, which can be tuned based on intuition or through experiments.

The approach here is different. Previous works use the model to generate candidate equations, but not to assess the appropriateness of the function, since typically the candidate with the highest likelihood (or lowest mean squared error) is selected. If, instead, one viewed the generation as a Monte Carlo sampling of the functional prior and marginalised results over the generated equations, then the two approaches would be more aligned. Moreover, we note that, for transformers, the training-set functions are generated randomly, with no context-specific knowledge (beyond the number of nodes, depth of tree, range of parameters, range of inputs or possible basis functions). Here we adopt a conceptually different approach, where the training set informs us which combinations of operators are \textit{a priori} more likely.
Of course, the hyperparameters which control the generation of the training set or the probabilities in the probabilistic grammar could be tuned to act as a prior over equations \citep{Becker_2022}. Prior knowledge has been previously incorporated into neural network based SR methods, provided that this knowledge can be expressed as a (nonlinear) inequality or equality \citep{Kubalik_2023}, which is a different scenario to that considered here. Finally, we note that our language model is designed not to have a limit on the number or range of the input variables; this can be a limitation of the transformer method because changing the input variables can require retraining \citep{Vastl_2022}.

In this work we consider the effects of accuracy, a prior on parameters and a prior on functions in model selection for SR. The first of these is always considered in SR (although the metric for it varies), whereas the latter two are less commonly used. In particular, \citet{Guimera_2020} introduce a prior on functions which assumes each node of the tree to be independent, but they do not account for the importance of the parameter prior when computing the evidence (which we argue in \cref{sec:Parameter priors} is important). \citet{Werner_22} overcome the issue of parameter priors (insofar as they choose a prior which can be normalised) by choosing Gaussian, zero-mean, isotropic priors for all parameters with a scalar hyperparameter which is optimised to maximise the marginal likelihood. One could question whether zero-mean or isotropic are good assumptions in general, and they assume all functions are equally likely \textit{a priori}. Similarly, \citet{Bomarito_2023} assume equal prior probability for all functions, but overcome the issue of priors by using the Fractional Bayes Factor (\cref{eq:FBF}). \citet{ESR_2022} implicitly consider all three of these concerns through an implementation of the MDL principle, where the parameter prior corresponds to Rissanen's universal code for the integers \citep{Rissanen_1983} and the functional prior depends on the number of nodes, unique operators, parameters and the values of constants (c.f. \cref{eq:MDL function prior}). This does not allow a general parameter prior, and its functional prior is uninformed by the frequency of occurence of combinations of operators in the domain of interest. 

\section{Conclusions}
\label{sec:Conclusions}

Model selection for SR trades off accuracy and complexity to obtain simple functions that describe a data set well. Previous attempts to do this have either used a heuristic metric, treated all functions as equally likely \textit{a priori} or neglected to construct a self-consistent parameter prior in a Bayesian framework. In the case where a structural prior on functions based on the frequency of operators in a corpus of equations had been used, the correlation between operators had been neglected. In this paper we attempt to reconcile these issues by formalising and investigating a range of model selection criteria based on the Bayesian evidence or Minimum Description Length principle. We explicitly compare the Bayesian and MDL procedures, implement the Fractional Bayes Factor to overcome the dependence of the Bayes factor on the arbitrary range of uninformative parameter priors and construct a Katz back-off language model on a set of scientific equations to learn a prior on operator ordering that can in principle take a function's entire tree into account. We compare our models with literature standards on benchmarks from the Nguyen and Korns datasets as well as a real-world test-case in the field of cosmology.
For standard SR benchmarks, we find that the MDL procedure without a prior on functions typically performs the best, however this method produced physically unrealistic equations on the astrophysical data. This is alleviated by applying the language model prior, suggesting this to be a promising avenue for SR modelling when domain-specific prior information is available.
We make the back-off model code and training equations publicly available
(\url{https://github.com/DeaglanBartlett/katz}).

\begin{acks}
We thank Alan Heavens, Florent Leclercq and Roberto Trotta for useful discussions. 
DJB is supported by the Simons Collaboration on ``Learning the Universe.''
HD is supported by a Royal Society University Research Fellowship (grant no. 211046).
PGF acknowledges support from European Research Council Grant No: 693024, STFC and the Beecroft Trust.
For the purpose of open access, we have applied a Creative Commons Attribution (CC BY) licence to any Author Accepted Manuscript version arising.
\end{acks}

\bibliographystyle{ACM-Reference-Format}
\bibliography{references}

\end{document}